\documentclass[
]{ceurart}

\usepackage{listings}
\usepackage{subfig}
\begin{document}

\copyrightyear{2021}
\copyrightclause{Copyright for this paper by its authors.
  Use permitted under Creative Commons License Attribution 4.0
  International (CC BY 4.0).}
% \conference{Woodstock'21: Symposium on the irreproducible science,
%   June 07--11, 2021, Woodstock, NY}
\conference{Forum for Information Retrieval Evaluation, December 13-17, 2021, India}

\title{Hate and Offensive Speech Detection in Hindi and Marathi}

\begin{center}
\author[1]{Abhishek Velankar}[%
email=velankarabhishek@gmail.com,
]

\author[1]{Hrushikesh Patil}[%
email=hrushi2900@gmail.com,
]
\author[1]{Amol Gore}[%
email=amolgore2512@gmail.com,
]
\author[1]{Shubham Salunke}[% 
email=shubhamsalunke30012001@gmail.com,
]
\address[1]{Pune Institute of Computer Technology, Pune, Maharashtra}

\author[2]{Raviraj Joshi}[%
email=ravirajoshi@gmail.com,
]
\address[2]{Indian Institute of Technology Madras, Chennai, Tamilnadu}
\end{center}

\begin{abstract}
Sentiment analysis is the most basic NLP task to determine the polarity of text data. There has been a significant amount of work in the area of multilingual text as well. Still hate and offensive speech detection faces a challenge due to inadequate availability of data, especially for Indian languages like Hindi and Marathi. In this work, we consider hate and offensive speech detection in Hindi and Marathi texts. The problem is formulated as a text classification task using the state of the art deep learning approaches. We explore different deep learning architectures like CNN, LSTM, and variations of BERT like multilingual BERT, IndicBERT, and monolingual RoBERTa. The basic models based on CNN and LSTM are augmented with fast text word embeddings. We use the HASOC 2021 Hindi and Marathi hate speech datasets to compare these algorithms. The Marathi dataset consists of binary labels and the Hindi dataset consists of binary as well as more-fine grained labels. We show that the transformer-based models perform the best and even the basic models along with FastText embeddings give a competitive performance. Moreover, with normal hyper-parameter tuning, the basic models perform better than BERT-based models on the fine-grained Hindi dataset. 
\end{abstract}

\begin{keywords}
  Natural Language Processing \sep
  Convolutional Neural Networks \sep
  Long Short Term Memory \sep
  FastText \sep
  BERT \sep
  Hate Speech Detection.
\end{keywords}

\maketitle
\section{Introduction}

Hate speech is often defined as the use of hateful language with the intent of attacking a person or a group to provoke, intimidate, express contempt or cause harm to them or on the basis of their race, religion, ethnic origin, disability or gender \cite{macavaney2019hate,matamoros2021racism}. The advancement of technology has led to an increase in the use of social media and its accessibility across the globe. Several online social media users post harmful content without realizing that their actions often offend a person or a group of people \cite{banko2020unified,jiang2021understanding}. It is therefore important to automatically detect and filter out such harmful content from the massive textual content being posted online every day \cite{schmidt2017survey,del2017hate,aluru2020deep}.

Hindi is one of the official languages of India and is spoken by around 45\% of its population \cite{Joshi_2020}. Due to its popularity in India, there are a large number of social media activities performed in the Hindi language written in Devanagari script. It is therefore important to detect hate speech in the Hindi language.

Marathi is the native language of Maharashtra state in India. It is spoken by around 83 million people all over the country and it ranks as the third most spoken language in India. People find it easier to express their opinions in regional languages and hence social media activities in Marathi have been quite popular among the Marathi-speaking diaspora. Most of the work in the area of sentiment analysis and hate speech detection has been concentrated on English \cite{kulkarni2021l3cubemahasent}. Exposure to native low-resource languages has been increasing in recent times. We specifically focus on low-resource Indian languages Hindi and Marathi.

In this work, we treat hate speech detection as a text classification problem and explore various deep learning approaches for the task. The datasets used are provided in the HASOC 2021 shared task \cite{hasoc2021mergeoverview}. These datasets consisted of text from different Twitter posts, tagged manually as hate and non-hate. The Marathi dataset has binary labels whereas the Hindi dataset consists of binary labels as well as more fine-grained labels namely none, hate, offensive, and profane. We analyze CNN and LSTM based models for the binary classification task. The word embeddings are initialized using corresponding Hindi or Marathi FastText word vectors. We also evaluated transformer-based models, particularly variations of BERT such as indicBERT, mBERT, RoBERTa for Hindi and Marathi \cite{devlin2018bert, kakwani2020inlpsuite}. A hierarchical approach is used for the fine-grained 4-class classification task in Hindi where we first distinguish the text between hate and non-hate class and use the text with hate class for further classification into three labels including HATE, OFFN, and PRFN. The hierarchical approach is compared with its direct multi-class counterpart. The best BERT models for each of the tasks are shared publicly\footnote{https://huggingface.co/l3cube-pune/hate-bert-hasoc-marathi}
\footnote{https://huggingface.co/l3cube-pune/hate-roberta-hasoc-hindi}
\footnote{https://huggingface.co/l3cube-pune/hate-multi-roberta-hasoc-hindi}.

\section{Related Work}

The low resource nature of Hindi and Marathi languages has limited the extent of work on hate speech detection in these languages. Typical deep learning models like CNN 1D, LSTM, and BiLSTM along with domain-specific word embeddings were evaluated on Hindi-English code mixed dataset in \cite{kamble2018hate}. They also showed that the above deep learning models perform way better than traditional machine learning approaches such as SVM, and random forests. 

A comparative study between machine learning and deep learning architectures for hate speech detection is proposed in \cite{dhamija2021comparative} where datasets containing English tweets have been used. Different combinations of feature engineering have been experimented which include machine learning models like Logistic Regression, Decision Trees, Random Forest, Naive Bayes, etc with TF-IDF and BOW vectorizers. Pre-trained embeddings GLoVe and custom word vectors have been used to train LSTM and GRU models. 

In \cite{Mujadia2019IIITHyderabadAH} authors compared different machine learning and neural network approaches for hate text speech detection in Hindi, with further classification in hate, offensive, and profane. The classical machine learning models like Linear SVM, Adaboost or Adaptive Boosting, Random Forests, Voting Classifier were used and LSTM based deep learning approaches were also used. They observed that machine learning models work better than neural network models in low-resource settings.

Various Hindi text classification approaches have been studied in \cite{Joshi_2020} using BOW, CNN, LSTM, BiLSTM, BERT, and LASER models. The work is particularly focused on Hindi text classification. It is shown that CNN with Hindi fast text embeddings performs the best. Additionally LASER has given very close results to the best performing model as compared to BERT. 

In \cite{Joshi_2021} authors propose approaches for hostile post detection in Hindi. Tests were performed on models like CNN, Multi-CNN, BiLSTM, CNN+BiLSTM, IndicBert, mBert along with FastText embedding provided by both IndicNLP and Facebook. This work shows that BERT-based models work slightly better than basic models. The multi CNN model with IndicNLP FastText word embedding performs best within the basic models.

\section{Dataset Details}

We used the hate speech detection datasets provided in HASOC 2021 shared task for Hindi and Marathi. The text for both datasets was obtained from Twitter.
\textbf{Marathi Dataset Description \cite{gaikwad2021cross}:} The dataset consisted of 1874 training samples with an average of 13 words per sentence. The class-wise details are shown in Table \ref{table:1}. It contained a total of 625 testing samples.
\begin{table*}[hbt!]
%   \caption{Marathi dataset label description and distribution}
\centering
\caption{Marathi dataset label description and distribution}
\label{table:1}
\begin{tabular}{c c c}
    % &Table 1: Marathi dataset label description and distribution&\\
    \midrule
    Category&Description&No. of Training Samples\\
    \midrule
    HOF  & Hate and Offensive content & 669\\
    NOT & Does not contain any hate, offensive, profane content & 1205\\
   \bottomrule
\end{tabular}
\end{table*}
\textbf{Hindi Dataset Description \cite{hasoc2021overview}:} The Hindi training dataset included a total of 4594 training samples which were divided into two tasks with an average of 26 words per sentence. Task 1 contained binary labels similar to Marathi i.e. HOF and NOT. Task 2 contained multiclass labels with 4 classes namely NONE, OFFN, HATE, PRFN. Even though labels in task 2 may sound similar, they are different by meaning as described in Table \ref{table:2}. A total of 1532 test samples was provided for both tasks.
% \clearpage
\begin{table}[hbt!]
%   \caption{Hindi dataset label description and distribution}
%   \label{tab:commands}
\centering
\caption{Hindi dataset label description and distribution}
\label{table:2}
  \begin{tabular}{c c c}
    % & Table 2 : Hindi dataset label description and distribution &\\
    \midrule
    & Task 1 & \\
    \bottomrule
    Category &A Description & No. of Training Samples\\
    \midrule
    HOF & Hate and Offensive content & 1433\\
    NOT & Does not contain any hate, offensive, profane content & 3161\\
    \bottomrule
    & Task 2 & \\
    \midrule
    NONE & Does not contain any hate or offensive content & 3161\\
    OFFN & The posts contain offensive language & 654\\
    HATE & Hate speech content & 566\\
    PRFN &Profane words are used & 213\\
    \bottomrule
  \end{tabular}
\end{table} 

\begin{figure}[hbt!]
  \centering
    \subfloat{\includegraphics[scale=0.5]{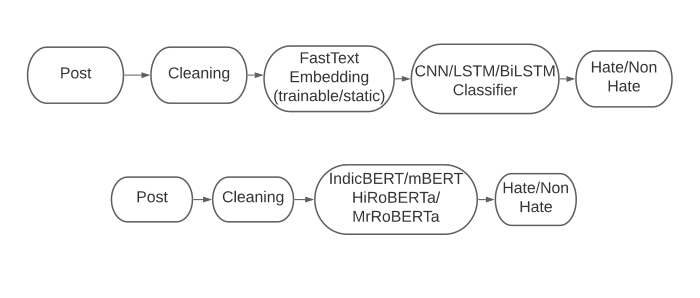}}
  \caption{Flow of binary classification}
  \label{fig:flow_diag}
\end{figure}

% \clearpage

\begin{figure}[hbt!]
  \centering
    \subfloat{\includegraphics[scale=0.5]{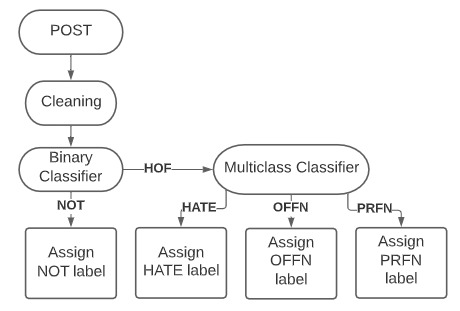}}
  \caption{Hierarchical approach representation}
  \label{fig:hie_diag}
\end{figure}

\section{Model Architectures}
We are using common deep learning text classification approaches for the task of Hate speech detection \cite{kulkarni2021experimental}. The models are used directly for binary classification tasks whereas a hierarchical approach is used for multi-labeled fine-grained classification. For each of the models, we selected the epoch giving maximum validation accuracy. We used the learning rate of 0.001 for CNN and LSTM based models whereas 5e-5 for BERT-based models. The general flow of the classification process is outlined in Figure \ref{fig:flow_diag} and Figure \ref{fig:hie_diag}. The models and the configurations are outlined below. \\ \\
\textbf{CNN:}
The basic CNN model used a 1D convolution layer with a filter of size 300 and kernel of size 3 with relu activation, followed by max-pooling with pool size 2, the same layers were added again, followed by 1D global max pooling. This is followed by a dense layer of size 50 and relu activation. Finally, the last layer with 2 nodes with softmax activation was used. Dropout of 0.3 was used after the 1D max-pooling layer.\\ \\
\textbf{LSTM:}
For the LSTM model LSTM layer with 32 nodes followed by 1D global max-pooling was used, then a dense layer with 16 nodes along with relu activation was used, followed by 0.2 dropout and finally, a dense layer with 2 nodes with softmax activation was used.\\ \\
\textbf{BiLSTM:}
Bi-LSTM layer with 300 nodes followed by 1D global max-pooling layer was used, then dense layer with 100 nodes along with activation relu was used. This was followed by a dropout of 0.2, then the final layer with 2 nodes with activation softmax was used.\\ \\
\textbf{BERT:}
BERT is a pre-trained language model on a large publicly available text corpus. It is a transformer-based model which is bi-directional in nature. It is pre-trained using two tasks- masked language modeling and next sentence prediction. We evaluated some variations of BERT for both Hindi and Marathi tasks \cite{wolf2020transformers}.
\begin{itemize}
    \item Multilingual BERT\footnote{https://huggingface.co/bert-base-multilingual-cased}: Pre-trained on 104 top languages worldwide including Hindi and Marathi.
    \item IndicBERT\footnote{https://huggingface.co/ai4bharat/indic-bert}: Pre-trained on 12 major Indian languages released by Ai4Bharat.
    \item roberta-base-mr\footnote{https://huggingface.co/flax-community/roberta-base-mr}: Released by flax-community, pre-trained on Marathi with masked language modeling objective.
    \item roberta-Hindi\footnote{https://huggingface.co/flax-community/roberta-hindi}: RoBERTa base model for Hindi released by flax-community.
    \item indic-transformers-hi-bert\footnote{https://huggingface.co/neuralspace-reverie/indic-transformers-hi-bert}:  BERT model pretrained on OSCAR corpus released by neuralspace-reverie.
\end{itemize}

\textbf{Hierarchical Approach:} 
\begin{itemize}
    \item The first model is trained on task 1 having binary labels HOF and NOT.
    \item The second model is trained on ternary labels defined in Task 2 by removing entries having NONE values. The ternary labels include OFFN, HATE, and PRFN. 
    \item The test data is passed through the first model to get the corresponding output labels as HOF or NOT.
    \item The samples predicted as HOF labels are further passed to the second model for classifying them into HATE, OFFN, and PREN labels, results from both models are then combined for the final result.
\end{itemize}

\begin{table}[h!bt!]
%   \label{tab:basic_res}
% \caption{\label{tab:freq}Simple models Evaluation Results.}
% \label{tab:freq}
\centering
\caption{Simple models Evaluation Results}
\label{table:3}
  \begin{tabular}{c c c c c c}
    % & &Table 3: Simple models Evaluation Results & &\\
    \midrule
    Model & Embedding & Accuracy & Macro & Macro &Macro \\
     &  & Score & F1 &Precision &Recall \\
    \midrule
    & & Marathi & &  \\
     \bottomrule
       & Random & 0.737 & 0.721 &0.854 &0.740 \\
     CNN& Trainable FastText &0.841 & 0.820 &0.877 &0.818 \\
     & Non-Trainable Fastext & 0.832 & 0.817 &\textbf{0.910} &0.832 \\
     \bottomrule
        & Random & 0.808 &0.782 &0.854 &0.782 \\
     LSTM& Trainable FastText &0.819  &0.795 &0.862 &0.794 \\
     & Non-Trainable Fastext &\textbf{0.859}  &\textbf{0.842} & 0.900&0.844 \\
    \bottomrule 
      & Random & 0.787 & 0.767 & 0.868 & 0.778 \\
     BiLSTM& Trainable FastText &0.803 & 0.777 &0.850 &0.776 \\
     & Non-Trainable Fastext & 0.854 & 0.839 &0.909 &\textbf{0.847} \\
    \bottomrule
    \toprule
    & & Hindi Task 1 & &  \\
    \midrule
       & Random & 0.735 & 0.701 &0.80 &0.701 \\
     CNN& Trainable FastText &0.780 & 0.740 &0.810 &0.730 \\
     & Non-Trainable Fastext & 0.780 & \textbf{0.760} &\textbf{0.850} &\textbf{0.760} \\
     \bottomrule
        & Random & 0.734 &0.703 &0.808 &0.705 \\
    LSTM & Trainable FastText &0.750  &0.710 &0.790 &0.700 \\
     & Non-Trainable Fastext &0.760  &0.750 & 0.830&0.750 \\
    \bottomrule 
      & Random & 0.751 &0.712 &0.802 & 0.708 \\
    BiLSTM & Trainable FastText &0.760 & 0.712 &0.781 &0.703 \\
     & Non-Trainable Fastext & \textbf{0.800} & 0.745 &0.796 &0.726 \\
     \bottomrule
     \toprule
    & & Hindi Task 2 & &\\
    \midrule
    & & Direct 4 Class & &\\
    \midrule
    CNN & Non-Trainable FastText & \textbf{0.753} & 0.549 & \textbf{0.618} & 0.517 \\
    \midrule
    & & Hierarchical Approach & &\\
    \midrule
    CNN & Non-Trainable FastText & 0.734 & \textbf{0.554} & 0.587 & \textbf{0.534} \\
    \bottomrule
  \end{tabular}
\end{table}

\begin{table}[h!bt!]
%   \label{tab:transformer_res}
% \caption{\label{tab:freq}Transformer-based Models Evaluation Results.}
\centering
\caption{Transformer-based Models Evaluation Results}
\label{table:4}
  \begin{tabular}{c c c c c}
    % & Table 4: Transformer-based Models Evaluation Results  &\\

    \midrule
    Model & Accuracy & Macro &Macro &Macro \\
     &  Score & F1 &Precision &Recall \\
     \midrule
     &  Marathi  & \\
     \midrule
      indicBERT & \textbf{0.880} & \textbf{0.869} &\textbf{0.871} &\textbf{0.867} \\
     
       mBERT & 0.860 &0.848 &0.844 &0.853 \\
    
     RoBERTa-Base-Mr & 0.870 & 0.850 & 0.858 & 0.844 \\
    \bottomrule
    
     &  Hindi Task 1  &  \\
      \midrule
      indicBERT & 0.770 & 0.720 &0.747 &0.708 \\
    
      RoBERTa Hi  & \textbf{0.800} &\textbf{0.763} &\textbf{0.778} &\textbf{0.754}\\
     
     Neural space BERT Hi & 0.761 & 0.687 & 0.746 & 0.674 \\
    \bottomrule
    
    &  Hindi Task 2 &   \\
    \midrule
    &  Direct 4 Class &   \\
    \midrule
    RoBERTa Hi & 0.711 & 0.521 &0.553 &0.499 \\
    \midrule
    &  Hierarchical Approach &   \\
    \midrule
    RoBERTa Hi & \textbf{0.724} & \textbf{0.540} &\textbf{0.567} &\textbf{0.520} \\
    \bottomrule
     
  \end{tabular}
\end{table}

\section{Results and Discussion}
\begin{figure}[hbt!]
  \centering
    \subfloat{\includegraphics[scale=0.45]{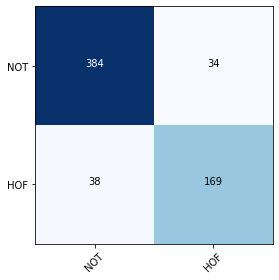}}
    % \hspace{2cm}
    \subfloat{\includegraphics[scale=0.45]{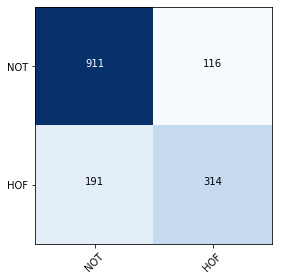}}
    % \hspace{2cm}
    \subfloat{\includegraphics[scale=0.45]{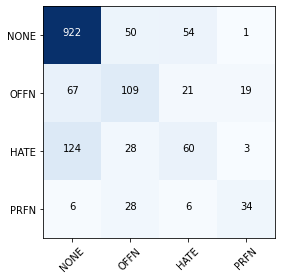}}
  \caption{Confusion Matrices for best models in Marathi binary (left), Hindi binary (middle) and Hindi multiclass (right) task respectively}
  \label{fig:confusion}
\end{figure}

In this work, the performance of different CNN and LSTM based models with and without FastText embeddings was evaluated on HASOC 2021 Marathi and Hindi datasets. Additionally, transformer-based models, particularly variations of BERT were used for comparison. 
Firstly, all three basic models CNN, LSTM, BiLSTM were trained with random word embedding initialization. The word embeddings were also initialized using pre-trained fast text embedding by IndicNLP and then used in trainable or static mode. The non-trainable fasttext embedding seems more promising than trainable fasttext and random embedding. In this case, the embeddings do not overfit the training data. The results of the basic models are described in Table \ref{table:3}. \\
Table \ref{table:4} summarizes the performances of different BERT models. It shows that indic BERT outperforms others in Marathi. For Hindi task 1, the RoBERTa Hindi base model performs the best. For Hindi Task 2, a hierarchical approach is used where two RoBERTa Hindi base models were trained, first for binary and second for ternary classification removing the NONE values. This model performs better than direct multiclass classification but slightly lower than FastText + CNN setting for Task 2. We observe that BERT models are more susceptible to data imbalance in Hindi fine-grained task and requires oversampling from underrepresented classes. Whereas basic models are more robust to such imbalance, the direct 4-way approach performs better than hierarchical classification. The confusion matrices for the best model in each task is shown in Figure \ref{fig:confusion}.

% \begin{figure}[hbt!]
%   \centering
%   \includegraphics[width=.5\linewidth]{mr_indicBert.png}
%   \caption{Confusion Matrix for Marathi Binary classification}
% \end{figure}

% \clearpage

% \begin{figure}[hbt!]
%   \centering
%   \includegraphics[width=.5\linewidth]{hi_roberta.png}
%   \caption{Confusion Matrix for Hindi Binary Classification}
% \end{figure}

% \begin{figure}[hbt!]
%   \centering
%   \frame{\includegraphics[width=.5\linewidth]{cnn_hie.png}}
%   \caption{Confusion Matrix for Hindi Multiclass Classification}
% \end{figure}

% \clearpage

\section{Conclusion}
In this work, we compared different deep learning approaches on Hindi and Marathi datasets from the HASOC 2021 shared task. The task included both binary and multiclass classification. For binary classification in Marathi and Hindi task 1, CNN and LSTM based models were used along with random and FastText embeddings. Out of these, the LSTM + non-trainable FastText setting worked the best for Marathi. In the case of Hindi, BiLSTM + non-trainable FastText performed better. Additionally, we experimented on different transformer-based BERT models like indicBERT, mBERT, RoBERTa-base for Marathi and RoBERTa base, and Neural space BERT for Hindi. IndicBERT outperformed other models for Marathi whereas RoBERTa performed the best for Hindi. The same RoBERTa model was used for the hierarchical approach. We show that transformer-based models perform better for binary tasks, but even basic models perform competitively. For Hindi task 2, it is shown that CNN + non-trainable FastText model performs slightly better than RoBERTa Hindi model.

\begin{acknowledgments}
This work was done under the L3Cube Pune mentorship
program. We would like to express our gratitude towards
our mentors at L3Cube for their continuous support and
encouragement.
\end{acknowledgments}

%%
%% Define the bibliography file to be used
\bibliography{main}

% \appendix

\end{document}